\documentclass{article}

\usepackage{multicol}
\usepackage[final,nonatbib]{tackling_climate_workshop_style}

\usepackage[utf8]{inputenc} % allow utf-8 input
\usepackage[T1]{fontenc}    % use 8-bit T1 fonts
\usepackage{hyperref}       % hyperlinks
\usepackage{url}            % simple URL typesetting
\usepackage{booktabs}       % professional-quality tables
\usepackage{amsfonts}       % blackboard math symbols
\usepackage{nicefrac}       % compact symbols for 1/2, etc.

\usepackage{graphicx, wrapfig}

\usepackage{microtype}      % microtypography
\usepackage[square,numbers]{natbib}
\title{Towards Global Crop Maps with Transfer Learning}

\author{%
Hyun-Woo Jo \thanks{Equal contribution.} $^{1}$ \quad Alkiviadis Koukos $^{*2}$ \quad Vasileios Sitokonstantinou$^2$ \\ 
\textbf{Woo-Kyun Lee}$^1$ \textbf{Charalampos Kontoes}$^2$ \\ 
$^1$ Department of Environmental Science and Ecological Engineering, Korea University \\
$^2$BEYOND Centre, IAASARS, National Observatory of Athens \\
\texttt{\{akoukos,vsito,kontoes\}@noa.gr}\\
\texttt{endeavor4a1@gmail.com}\\
\texttt{leewk@korea.ac.kr}
}

\begin{document}

\maketitle

\begin{abstract}
The continuous increase in global population and the impact of climate change on crop production are expected to affect the food sector significantly. In this context, there is need for timely, large-scale and precise mapping of crops for evidence-based decision making. A key enabler towards this direction are new satellite missions that freely offer big remote sensing data of high spatio-temporal resolution and global coverage. During the previous decade and because of this surge of big Earth observations, deep learning methods have dominated the remote sensing and crop mapping literature. Nevertheless, deep learning models require large amounts of annotated data that are scarce and hard-to-acquire. To address this problem, transfer learning methods can be used to exploit available annotations and enable crop mapping for other regions, crop types and years of inspection. In this work, we have developed and trained a deep learning model for paddy rice detection in South Korea using Sentinel-1 VH time-series. We then fine-tune the model for i) paddy rice detection in France and Spain and ii) barley detection in the Netherlands. Additionally, we propose a modification in the pre-trained weights in order to incorporate extra input features (Sentinel-1 VV). Our approach shows excellent performance when transferring in different areas for the same crop type and rather promising results when transferring in a different area and crop type.
\end{abstract}

\section{Introduction}

Food security, but also social and economic development are at high risk due to the population growth and climate change and the pressure they put on agriculture. Several recent studies indicate that the changes in climate cause substantial yield losses at the global level \cite{tito2018global, sultan2019evidence}. At the same time, the production should be increased by 24\% until 2030 - compared to 2022 - to achieve zero hunger, while reducing emissions by 6\% \cite{agr_outlook_2022}. 
% while the global food production in 2050 should be 60\%-110\% higher than that of 2005/2007 \cite{agr_outlook_2012, alexandratos2012world, ray2013yield}. 
Apart from climate-friendly policies and practices to ensure and promote the increase of agricultural productivity \cite{malhi2021impact}, there is also a need of global, timely and precise crop type mapping systems to assist in the monitoring and management of agricultural fields \cite{segarra2020remote}, the prediction of crop production \cite{van2020crop, khaki2019crop} and the effective spatial allocation of the agricultural practices \cite{folberth2020global}. 

Earth Observation (EO) data have been extensively used to train Machine Learning (ML) and Deep Learning (DL) models to produce crop maps \cite{d2021parcel, sitokonstantinou2021scalable, van2018synergistic, defourny2019near, du2019smallholder, you202110, jo2020deep}.
Nevertheless, most approaches require large labeled datasets for training
\cite{russwurm2019breizhcrops,m2019semantic, weikmann2021timesen2crop}.
% \cite{large_label_01, large_label_02},
In reality, ground samples that capture spatio-temporal differences of crops worldwide are scarce and this remains one of the main barriers for global applications \cite{label_barrier_01}, limiting most studies to small or homogeneous areas. The most trustworthy way to acquire such ground observations is field surveys that are time-consuming, expensive and cannot cover every part of the world (e.g. inaccessible/remote areas). 
Transfer Learning (TL) \cite{pan2009survey} has been successfully applied to overcome this issue, by improving the learning performance on reduced datasets while decreasing the computational complexity \cite{zhuang2020comprehensive}. To this direction, a handful of works have been published the past couple of years that apply TL for crop classification using EO data \cite{hao2020transfer,bosilj2020transfer,nowakowski2021crop,zhang2020mapping}.

In this paper, we apply TL on paddy rice mapping using only Sentinel-1 time series, by transferring knowledge from South Korea to European areas, for which there is a small amount of ground samples. We implement and apply a recurrent U-net model to detect paddy in South Korea using Sentinel-1 VH backscatter time-series as input. Then, we explore the capability of transferring the knowledge captured from the paddy rice model to effectively predict i) paddy rice in Spain and France and ii) summer barley in the Netherlands. Moreover, we explore fine-tuning the model by augmenting the input space with Sentinel-1 VV backscatter coefficients. 
 
% \section{Methods}
\section{Data \& Problem Formulation}
\paragraph{Data.}
We used Synthetic Aperture Radar (SAR) Sentinel-1 data and computed the 20-day-mean backscattering coefficient (VH|VV) for each pixel for a time-series throughout the growing period. Then we extracted patches of 256x256 pixels for the areas of interest, using Google Earth Engine. All input data were scaled using max-min normalization. The datasets of rice in South Korea (2017-2019), Spain (2021), and France (2020) consist of 12,942, 88, and 134 patches, respectively. The dataset of summer barley in Netherlands (2018) includes 2,280 patches, however with few barley pixels in the patches. In each dataset, 60\% of randomly selected patches were used for training and the rest 40\% were used for testing.

\paragraph{Problem Formulation.}
The time series of EO data are denoted by \(x_{r,t}^s\) and the annotated crop data by \(y^c_r\) where \(r, t, s, c\) represent region, time, feature, and crop type, respectively. In this study, the crop mapping was performed with a recurrent U-net model \((h)\) exploiting time-series during the growing period:
\begin{equation}
  \hat{y}_{r,t} = h(x_{r,1},x_{r,2},...,x_{r,t})
\end{equation} \label{eq1}
where \(t = 1...8\) indicates the relative time instance, i.e., 20-day feature vector, within the cultivation period. The model was pre-trained \((h^p)\) to classify paddy rice in South Korea by using Sentinel-1 VH backscatter, in patches extended all over the country and for the years of 2017-2019. However, unlike South Korea where rice is a staple food, the number of \(x\) in the other cases of \(r, c\) hardly suffice to efficiently train deep neural networks.

\section{Methodology}

\paragraph{3D Recurrent U-net.}
A custom recurrent U-Net (Fig. \ref{fig:runet}) was designed to exploit both spatial and temporal context of the EO time-series in order to produce timely paddy rice segmentation maps. The model follows a standard U-Net architecture; the encoder consists of a series of recurrent modules including convolutional layers, drop-out, and spatial max-pooling. In the recurrent module, each time step shares the convolution layers, and the weighted output of the previous time step is added to the next time step; thus, the phenological context can be passed to the later calculation. Considering both the preservation of temporal features and computational efficiency, the max-pooling layers at the skip connections were applied to the time axis so that the adjacent time steps at the same developing phase were pooled, and half the size of the features were passed to the decoder.

\paragraph{Transfer Learning.}

We implement different scenarios of TL to identify an optimal application according to data availability and similarity. As a baseline, we transferred only the architecture with randomly initialized weights ($RI$). The others include initializing with the pre-trained weights and then fine-tuning \((f^{r,c,s}(\cdot))\) to adapt to the target \(r, c, s\). Considering that the model \((h)\) consists of an encoder \((h_E)\), which extracts the crop's phenological characteristics, and it is followed by a decoder \((h_D)\), the applications were to fine-tune the entire networks ($FT$), fine-tune only \(h_E\) while freezing \(h_D\) ($FT_E$) or fine-tune \(h_D\) while freezing \(h_E\) ($FT_D$). 

\begin{equation}
    RI = f(h_E \cdot h_D) 
    \quad
    FT =  f(h^p_E \cdot h^p_D)
    \quad
    FT_E = f(h^p_E) \cdot h^p_D
    \quad
    FT_D =  h^p_E \cdot f(h^p_D)
\end{equation} 
  
\paragraph{Incorporation of additional feature types.}
In crop classification, diverse characteristics of each crop (e.g., texture, reflection) raise the need of an extended application of TL, such as using different sources of data as input. In this direction, we adapt \(h^p\), pre-trained on Sentinel-1 VH backscatter, to take as input both Sentinel-1 VH and VV features. To do this, the pre-trained weights at the first layer of the encoder \((W^P_{E_0})\) are divided by the total number of input layers (Eq.\ref{eq3}). Therefore, a similar scale of signal intensity is transferred to the activation functions \((\sigma)\) that is invariant to the number of inputs, and ensures that \(h^p\) maintains the trained feature extraction process.
\begin{equation}
h^P_{E_0} = \sigma((W^P_{E_0} \cdot x^{s} + W^P_{E_0} \cdot x^{s'})/2 + b)
\end{equation} \label{eq3}

\section{Experiments and Results}
\paragraph{Experiments.} We implemented 10 scenarios by combining different \(r, c, s\), where fine-tuning was conducted in \(r_1\) and performance was tested in \(r_2\). The main goal of this study is to investigate the effect of TL for the same target labels (e.g., paddy rice) in different areas. Therefore, we run the following experiments: \(r_1\)-\(r_2\)-\(c\)-\(s\): 1) Spain-Spain-rice-VH, 2) Spain-Spain-rice-VH|VV, 3) Spain-France-rice-VH, 4) Spain-France-rice-VH|VV, 5) France-France-rice-VH, 6) France-France-rice-VH|VV), 7) France-Spain-rice-VH, 8) France-Spain-rice-VH|VV. Additionally, we explore the efficiency of TL in different regions and different crop types (summer barley in the Netherlands) to examine if the knowledge of paddy rice could contribute in mapping other crops; \(r_1\)-\(r_2\)-\(c\)-\(s\): 9) Netherlands-Netherlands-summer barley-VH, 10) Netherlands-Netherlands-summer barley-VH|VV. It is worth mentioning that the data of each region have been acquired from different years, which makes the application of TL even more challenging.

Based on our experiments, we found $FT_E$ achieved better performance than $FT$ and $FT_D$. Fine-tuning the decoder did not converge, whereas by fine-tuning only its last (or 2-3 last) layers the model was successfully trained, but provided suboptimal performance. Table \ref{tab:iou} presents the Intersection over Union (IoU) of the positive class, for the $RI$, $FT$ and $FT_E$ and the 10 different scenarios mentioned earlier. We also compare their performance against locally trained Random Forest (RF) models. Visual maps of predictions for each scenario and method can be found in the Appendix (Figures \ref{fig:spa_fra_vh}-\ref{fig:net_net_vhvv})

\begin{table}[!ht]
\centering
\caption{Mean IoU for the different scenarios and methods}\label{tab:iou}
\resizebox{\textwidth}{!}{\begin{tabular}{lcccccccccc}
\toprule
Fine-tuning & \multicolumn{4}{c}{Spain}  & \multicolumn{4}{c}{France} & \multicolumn{2}{c}{The Netherlands} \\ 
\midrule
Test & \multicolumn{2}{c}{Spain} & \multicolumn{2}{c}{France}   & \multicolumn{2}{c}{France} & \multicolumn{2}{c}{Spain} & \multicolumn{2}{c}{The Netherlands} \\ 
\midrule
Feature & VH & VH|VV & VH & VH|VV & VH & VH|VV & VH & VH|VV & VH & VH|VV \\ 
\midrule
RF   & 0.87 & \textbf{0.90} & \textbf{0.63} & \textbf{0.66} & 0.76 & 0.84  & 0.77 & 0.78 & 0.26 & 0.40  \\ 
RI   & 0.86 & 0.69 & 0.52 & 0.36 & 0.76 & 0.74 & 0.70 & 0.73 & 0.31 & 0  \\ 
FT   & 0.89 & \textbf{0.90} & 0.57 & 0.63 & 0.82 & 0.83 & \textbf{0.82} & 0.83 & 0.40 & 0.45  \\ 
FT\textsubscript{E}   & \textbf{0.90} & \textbf{0.90} & \textbf{0.63} & \textbf{0.66} & \textbf{0.86} & \textbf{0.86} & 0.79 & \textbf{0.84} & \textbf{0.42} & \textbf{0.54}  \\ 
\bottomrule
\end{tabular}}
\end{table}

Using only the model's architecture without transferring the pre-trained weights ($RI$), we observe a poor performance in most cases, which is even poorer in the case of VH|VV input. This can be explained by the fact that augmenting the input with more layers results in more parameters which, in combination with the few labels and no transferred knowledge, prevents the model from learning. On the other hand, fine-tuning the pre-trained U-net works very well for paddy rice mapping, both in the case of freezing the decoder and updating only the parameters of the encoder ($FT_E$) and in the case of updating the whole network's weights ($FT$). As expected, when fine-tuning and testing in the same area the performance is better. But when we fine-tune in Spain and test in France, we notice a significant drop in the IoU. 

\begin{wrapfigure}{r}{8cm}
\includegraphics[scale=0.25]{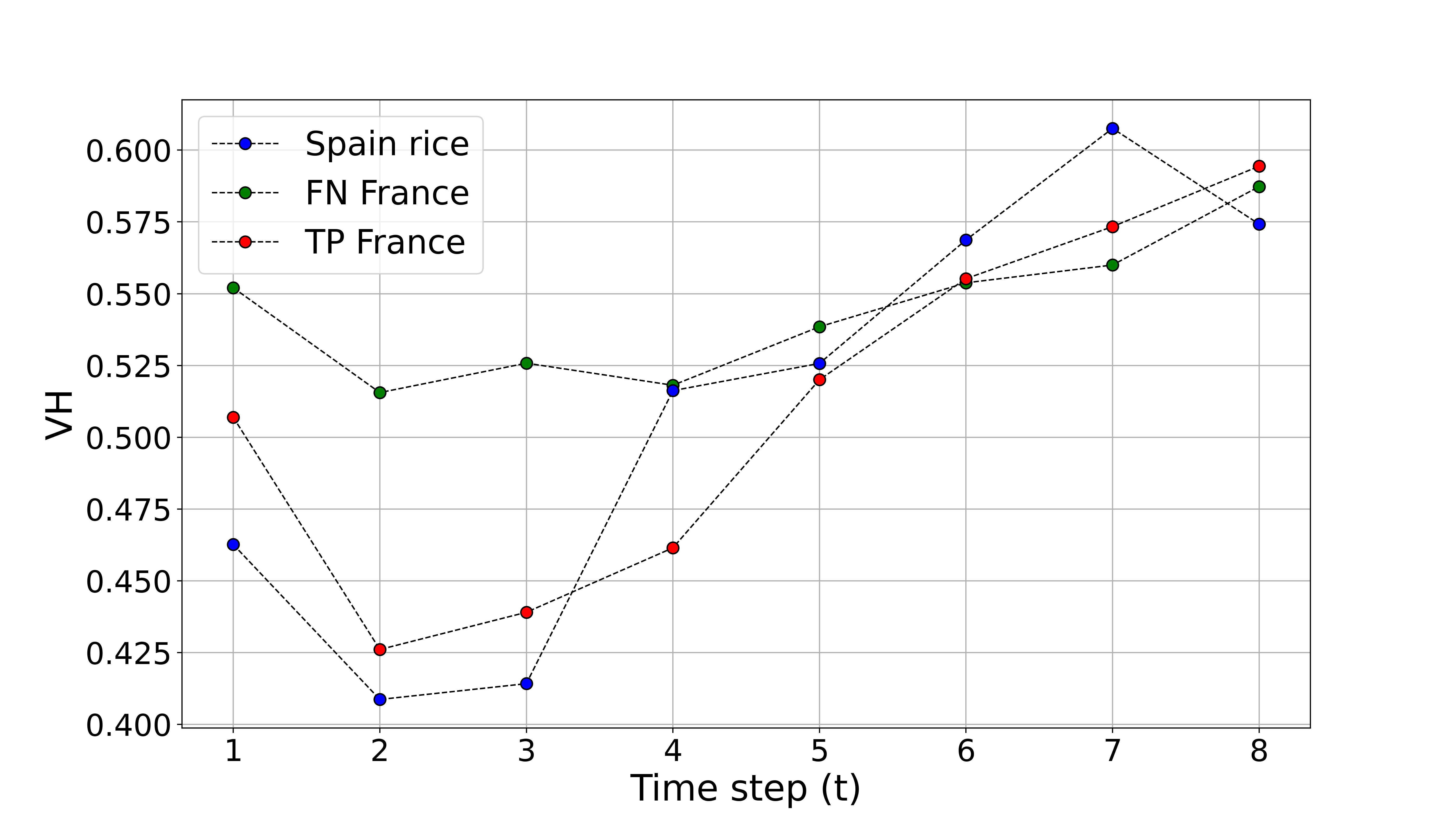}
\caption{Green and red dots represent the mean VH time-series of the True Positive (TP) and False Negative (FN) predictions of the reccurent U-net fine-tuned in Spain and tested in France. The blue dots represent the mean VH time-series of rice instances in Spain}\label{fig:vh}
\end{wrapfigure}

Figure \ref{fig:vh} shows the mean VH time-series of the True Positive (TP) and the False Negative (TN) of the predictions of the model fine-tuned in Spain and tested in France, together with the corresponding mean VH time-series of rice pixels in Spain.
TP (i.e., correctly predicted rice pixels) have very similar VH signature with that of the rice in Spain, whereas for FN predictions (i.e., rice instances that the model failed to identify) backscatter coefficients differ significantly, as compared to both the TP and the Spain rice pixels.
It seems that in France there are two different types of rice, out of which only one shares the same growing phases with those in Spain, and therefore the model is not able to successfully predict both of them.  
Moreover, the incorporation of the VV feature, especially in France, enhances slightly the performance. This is a strong indicator that TL not only works with the augmentation of the input with more features, but it can also provide better results.

Transferring the paddy rice model to predict summer barley does not perform as well. Paddy rice fields are intentionally flooded at the start of the cultivation period; SAR data have a great ability of identifying water content, which makes them ideal in classifying paddy rice. However, this is not the case for summer barley, and thus the discrimination of it using SAR data is much more challenging. Nevertheless, using both VH and VV backscatter coefficients we are able to achieve an IoU of 0.54 - also recall, precision and f1-score of 0.7  (Tables \ref{tab:recall}, \ref{tab:precision} and  \ref{tab:f1}) - which is interestingly high given the nature of the problem. In this case, we notice not only an improvement by using the extra input of VV, but rather a significantly better performance by fine-tuning only the encoder. 

It is also worth mentioning that RF also achieves great performance in almost every paddy rice experiment. However, it fails in predicting summer barley, even in the case of using both VV and VH, with an IoU of 0.4. Identifying paddy rice using EO data is not a particularly difficult problem, thanks to the flooding in the early vegetation period that was mentioned above. On the contrary, prediction of summer barley is a much more complex task, since it could share common phenological characteristics with other summer crop types (e.g., maize, summer wheat).

\section{Conclusion}
Precise, dynamic and detailed global crop type maps are essential for monitoring crop production that is under pressure. Such maps are powerful datasets that enable the timely identification of food security challenges and the large-scale, yet local-specific, rural planning to mitigate climate change. In this context, we propose a transfer learning method that leverages a pre-trained recurrent U-net model for paddy rice mapping in South Korea and fine-tunes it in other areas (France, Spain and the Netherlands) and/or crop types (paddy rice and summer barley) with a few available annotated data. TL for paddy rice mapping yielded excellent results both in Spain and in France. Based on our experiments, fine-tuning the encoder or the entire network provided the best performance, whereas fine-tuning the decoder did not converge. Additionally, the incorporation of an additional feature (i.e., VV backscatter coefficient) boosts the performance in almost every scenario. Finally, TL for barley in the Netherlands exhibits promising results, especially in the case of fine-tuning the encoder and incorporating the VV input, which outperforms significantly the corresponding RF model. 

\begin{ack}
This work was supported by the International Research and Development Program of the National Research Foundation of Korea (NRF) funded by the Ministry of Science and ICT [2021K1A3A1A78097879] and by the CALLISTO project, funded by EU's Horizon 2020 research and innovation programme under grant agreement No. 101004152.

\end{ack}

\newpage
% \section*{References}
\medskip

\bibliography{ref}

\newpage
\appendix{
\section{Supplemental Material}
\subsection{Model Details}

\begin{figure}[!ht]
    \centering
    \includegraphics[scale=0.8]{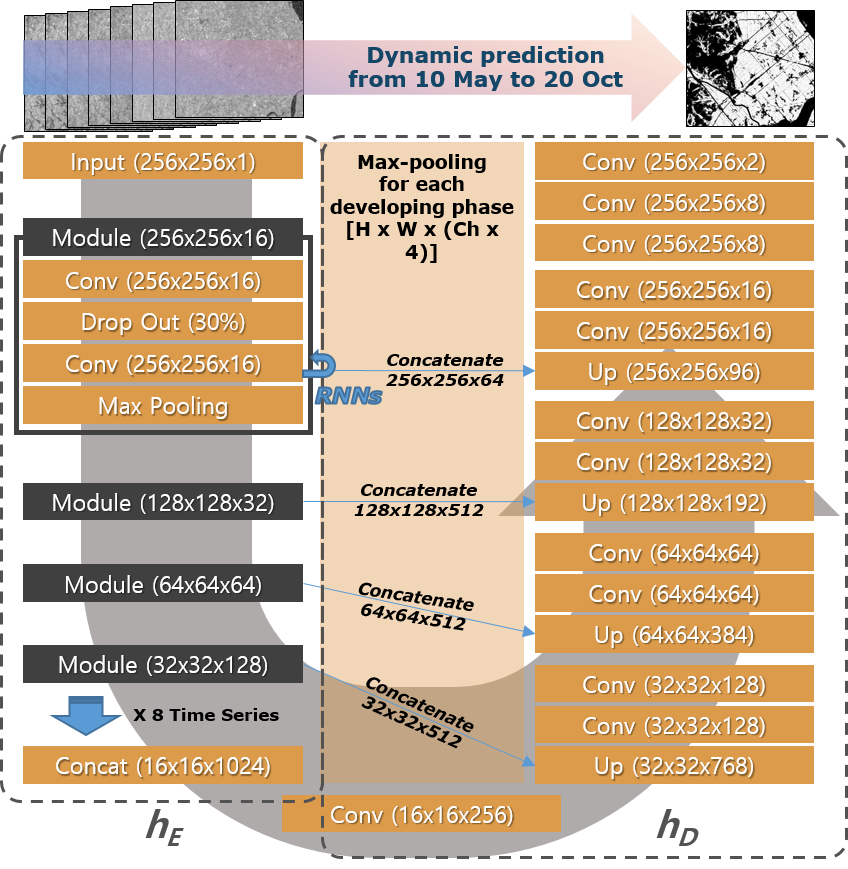}
    \caption{Recurrent U-net architecture}
    \label{fig:runet}
\end{figure}

\subsection{Training Details}

Apart from making predictions with the entire time-series, we are able to acquire predictions early in the cultivation period using zero padding in the later time steps (Figure \ref{fig:forgetting}). The prediction using the full time-series (\(t=8)\) and the early prediction using confined time steps (\(1\leq t<8)\) share a common feature extraction process but the signal intensity through the neural network can be greatly differed by \(\sigma\). Therefore, if the input of the aforementioned time-series is provided in a random order during the training phase, the loss function is hardly optimized and overfitted to lastly seen training instances, which is associated with the problem of catastrophic forgetting \cite{CF_problem}. 

Therefore, we manipulate the training order as described in Fig. \ref{fig:forgetting}, so that the model will be able to sequentially learn from each time-step's data. Specifically, in each epoch we firstly provide as input the cases of the very early prediction (e.g. only 1 time step) and lastly the ones of the full growing season. By training the model with chronologically ordered batches the parameters are updated gradually, with an additional time step after every new batch type, which is more likely to preserve the knowledge gained from the values of each new time step. 

\begin{figure}[!ht]
    \centering
    \includegraphics[scale=0.6]{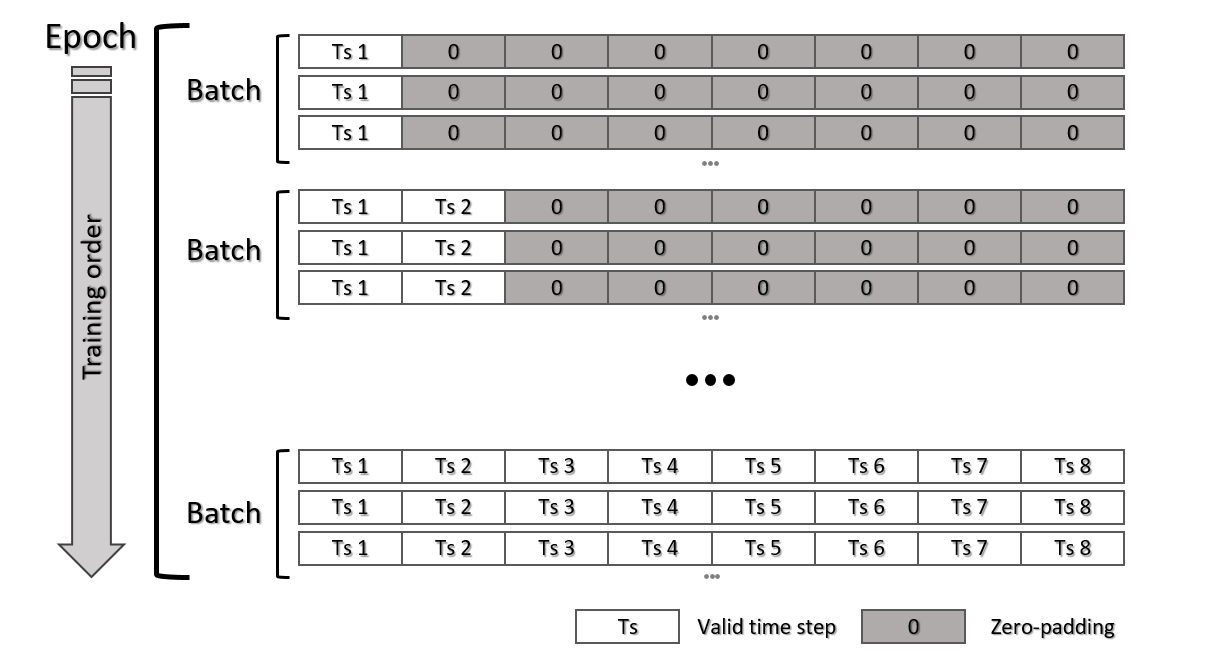}
    \caption{Chronologically ordered batch training}
    \label{fig:forgetting}
\end{figure}

\subsection{Additional results and plots}

\paragraph{Additional metrics}
Table \ref{tab:recall}, \ref{tab:precision} and \ref{tab:f1} present the recall, the precision  and the f1-score of the positive class for each of the different experiments, respectively. 

\begin{table}[!ht]
\centering
\caption{Recall for the positive class in the different scenarios and methods}\label{tab:recall}
\resizebox{\textwidth}{!}{\begin{tabular}{lcccccccccc}
\toprule
Fine-tuning & \multicolumn{4}{c}{Spain}  & \multicolumn{4}{c}{France} & \multicolumn{2}{c}{The Netherlands} \\ 
\midrule
Test & \multicolumn{2}{c}{Spain} & \multicolumn{2}{c}{France}   & \multicolumn{2}{c}{France} & \multicolumn{2}{c}{Spain} & \multicolumn{2}{c}{The Netherlands} \\ 
\midrule
Feature & VH & VH|VV & VH & VH|VV & VH & VH|VV & VH  & VH|VV & VH & VH|VV \\ 
\midrule
RF   & 0.957 & \textbf{0.964} & 0.665 & \textbf{0.698} & 0.831 & 0.881 & 0.878 & 0.846 & 0.293 & 0.446  \\ 
RI   & \textbf{0.967} & 0.765 & 0.584 & 0.460 & 0.879 & 0.841  & 0.809 & 0.818 & 0.400 & 0.004 \\
FT   & 0.957 & 0.962 & 0.601 & 0.674 & 0.896 & 0.894 & \textbf{0.954} & 0.950 & 0.490 & 0.614 \\
FT\textsubscript{E} & 0.964 & 0.962 & \textbf{0.674} & 0.694 & \textbf{0.915} & \textbf{0.911} & 0.914 & \textbf{0.964} & \textbf{0.518} & \textbf{0.705} \\
\bottomrule
\end{tabular}}
\end{table}

\begin{table}[!ht]
\centering
\caption{Precision for the positive class in the different scenarios and methods}\label{tab:precision}
\resizebox{\textwidth}{!}{\begin{tabular}{lcccccccccc}
\toprule
Fine-tuning & \multicolumn{4}{c}{Spain}  & \multicolumn{4}{c}{France} & \multicolumn{2}{c}{The Netherlands} \\ 
\midrule
Test & \multicolumn{2}{c}{Spain} & \multicolumn{2}{c}{France}   & \multicolumn{2}{c}{France} & \multicolumn{2}{c}{Spain} & \multicolumn{2}{c}{The Netherlands} \\ 
\midrule
Feature & VH & VH|VV & VH & VH|VV & VH & VH|VV & VH  & VH|VV & VH & VH|VV \\ 
\midrule
RF   & 0.910 & 0.926 & \textbf{0.926} & \textbf{0.931} & 0.901 & \textbf{0.943} & \textbf{0.866} & \textbf{0.906} & \textbf{0.718} & \textbf{0.781}  \\ 
RI   & 0.891 & 0.875 & 0.826 & 0.638 & 0.849 & 0.856 & 0.838 & 0.871 & 0.590 & 0.477 \\
FT   & \textbf{0.932} & 0.928 & 0.919 & 0.915 & 0.905 & 0.920 & 0.858 & 0.872 & 0.686 & 0.631 \\
FT\textsubscript{E} & 0.929 & \textbf{0.935} & 0.903 & \textbf{0.929} & \textbf{0.933} & 0.939 & 0.854 & 0.873 & 0.683 & 0.696 \\
\bottomrule
\end{tabular}}
\end{table}

\begin{table}[!ht]
\centering
\caption{F1-score of the positive class for the different scenarios and methods}\label{tab:f1}
\resizebox{\textwidth}{!}{\begin{tabular}{lcccccccccc}
\toprule
Fine-tuning & \multicolumn{4}{c}{Spain}  & \multicolumn{4}{c}{France} & \multicolumn{2}{c}{The Netherlands} \\ 
\midrule
Test & \multicolumn{2}{c}{Spain} & \multicolumn{2}{c}{France}   & \multicolumn{2}{c}{France} & \multicolumn{2}{c}{Spain} & \multicolumn{2}{c}{The Netherlands} \\ 
\midrule
Feature & VH & VH|VV & VH & VH|VV & VH & VH|VV & VH  & VH|VV & VH & VH|VV \\ 
\midrule
RF & 0.932 & 0.945 & \textbf{0.774} & \textbf{0.798} & 0.864 & 0.911 & 0.872 & 0.875 & 0.416 & 0.567  \\ 
RI   & 0.927 & 0.816 & 0.684 & 0.534 & 0.864 & 0.848 & 0.824 & 0.844 & 0.477 & 0.007 \\
FT   & 0.944 & 0.945 & 0.726 & 0.776 & 0.901 & 0.907 & \textbf{0.903} & 0.909 & 0.571 & 0.622 \\
FT\textsubscript{E} & \textbf{0.947} & \textbf{0.948}  & 0.772 & 0.794 & \textbf{0.924} & \textbf{0.925} & 0.883 & \textbf{0.916} & \textbf{0.589} & \textbf{0.700} \\
\bottomrule
\end{tabular}}
\end{table}

Below we present visual predictions of the different scenarios and for the different methods of crop mapping. In every Figure below, the first column illustrates a composited of the first three Sentinel-1 VH images (Image), the second the ground truth labels (Label), the third the RF predictions, the fourth the $RI$ predictions, the fifth the $FT$ predictions and the sixth the $FT_E$ predictions. 

\begin{figure}[!ht]
    \centering
    \includegraphics[scale=0.375]{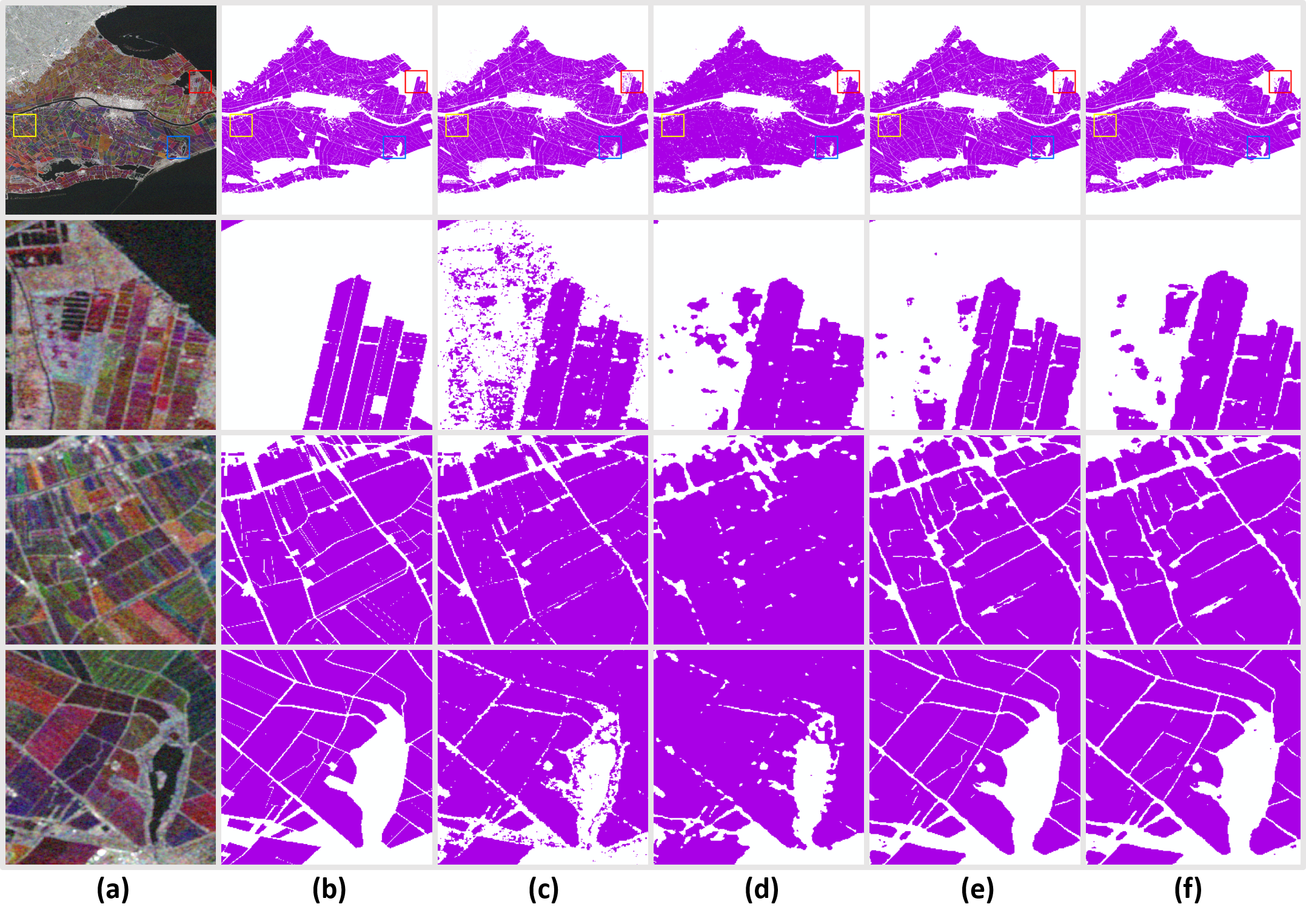}
    \caption{Visual comparisons on experiment 1 (Spain-Spain-rice-VH). The first row shows the overall results of the test image and the following rows show three randomly selected test patches. (a) Image. (b) Label. (c) RF. (d) RI. (e) FT. (f) FT\textsubscript{E}. }
    \label{fig:spa_spa_vh}
\end{figure}

\begin{figure}[!ht]
    \centering
    \includegraphics[scale=0.375]{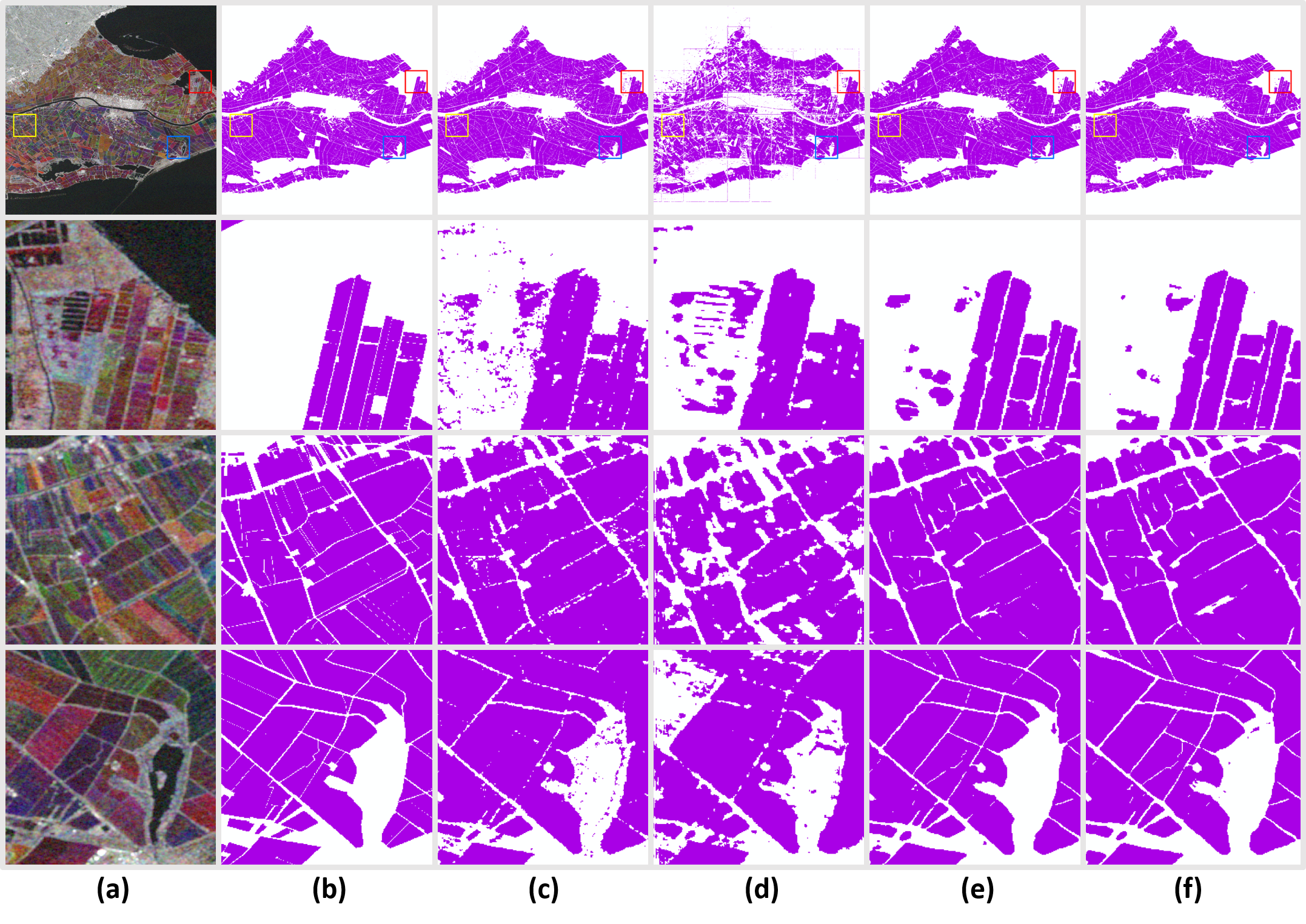}
    \caption{Visual comparisons on experiment 2 (Spain-Spain-rice-VH|VV). The first row shows the overall results of the test image and the following rows show three randomly selected test patches. (a) Image. (b) Label. (c) RF. (d) RI. (e) FT. (f) FT\textsubscript{E}. }
    \label{fig:spa_spa_vhvv}
\end{figure}

\begin{figure}[!ht]
    \centering
    \includegraphics[scale=0.375]{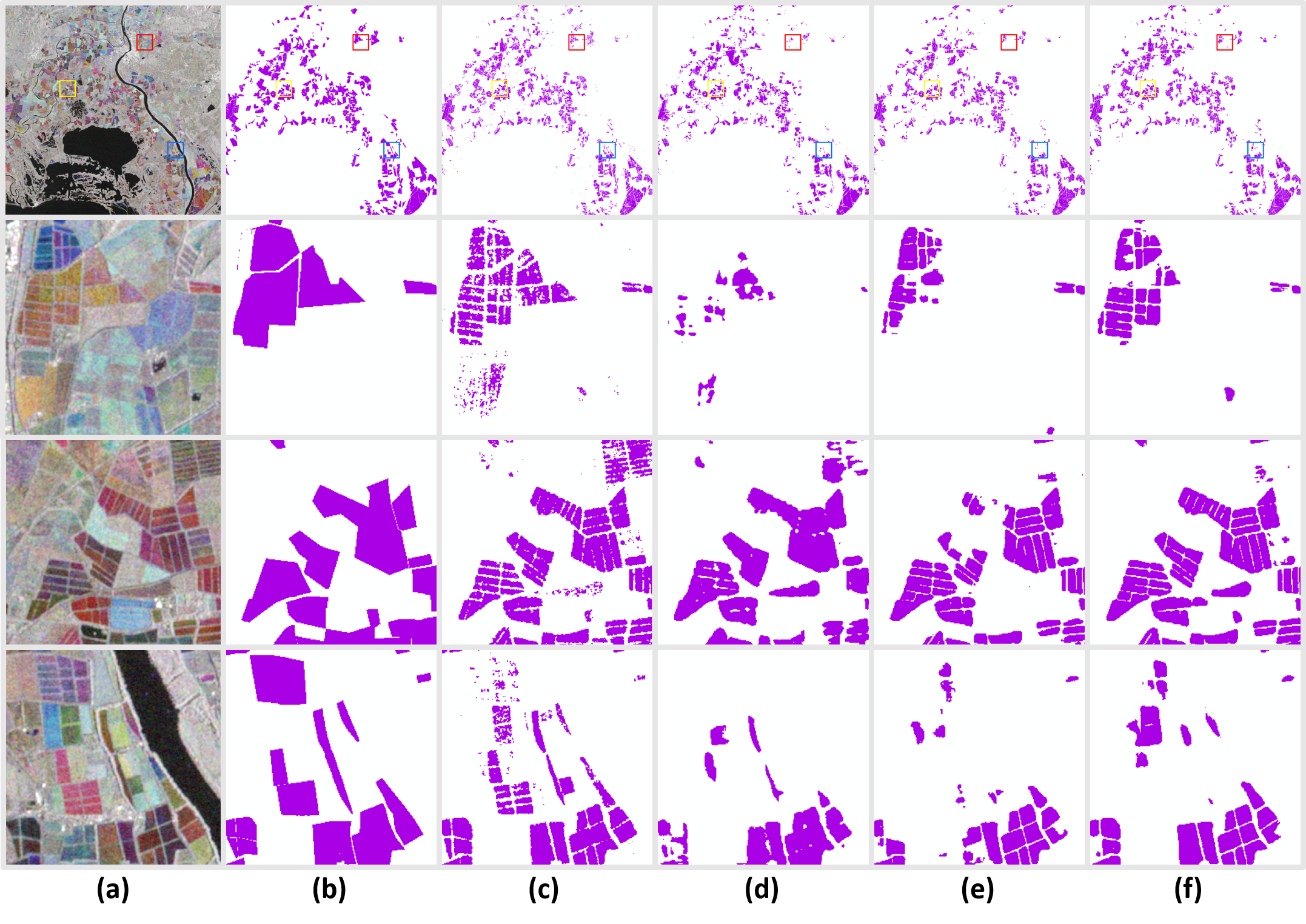}
    \caption{Visual comparisons on experiment 3 (Spain-France-rice-VH). The first row shows the overall results of the test image and the following rows show three randomly selected test patches. (a) Image. (b) Label. (c) RF. (d) RI. (e) FT. (f) FT\textsubscript{E}. }
    \label{fig:spa_fra_vh}
\end{figure}

\begin{figure}[!ht]
    \centering
    \includegraphics[scale=0.375]{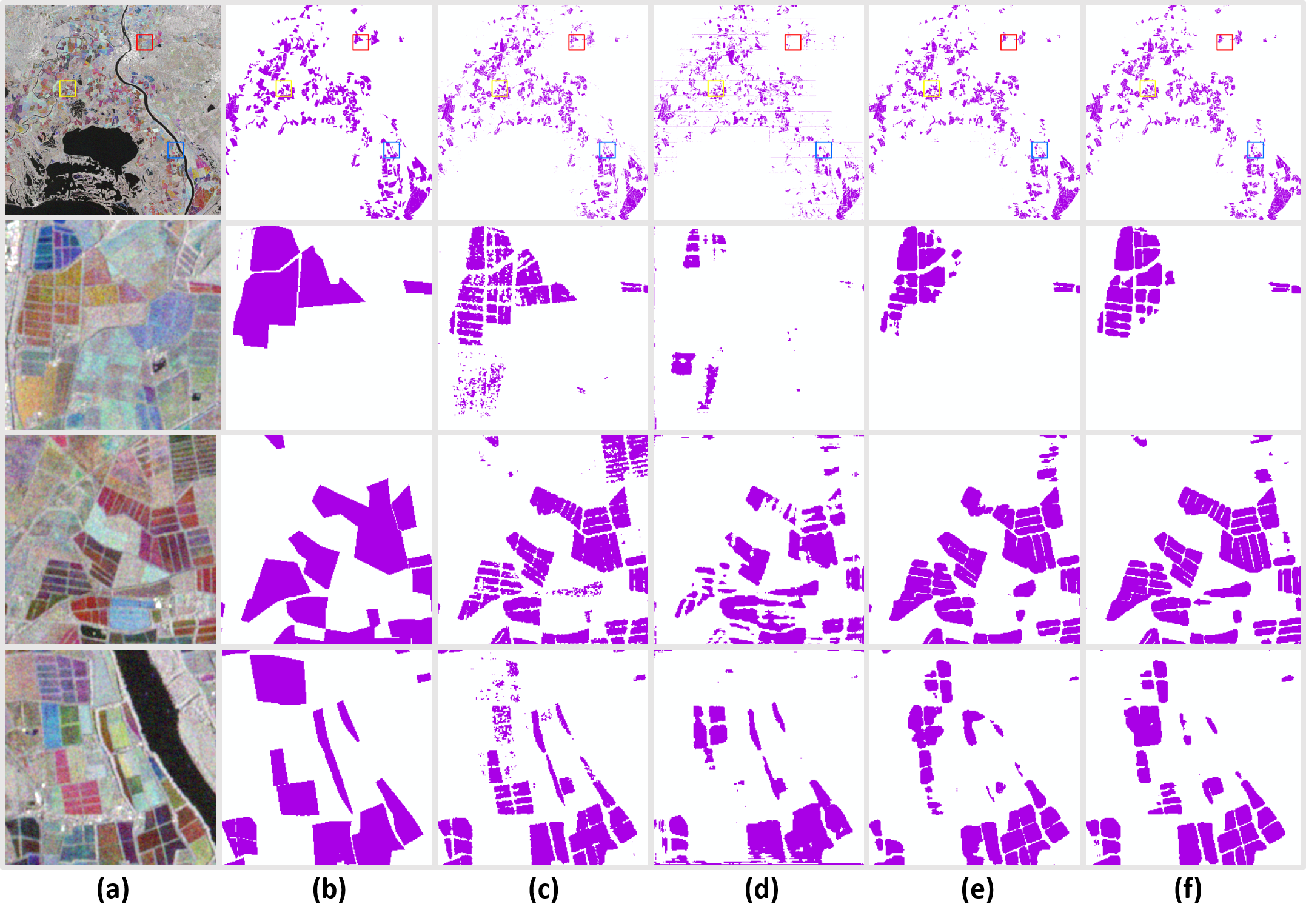}
    \caption{Visual comparisons on experiment 4 (Spain-France-rice-VH|VV). The first row shows the overall results of the test image and the following rows show three randomly selected test patches. (a) Image. (b) Label. (c) RF. (d) RI. (e) FT. (f) FT\textsubscript{E}. }
    \label{fig:spa_fra_vhvv}
\end{figure}

\begin{figure}[!ht]
    \centering
    \includegraphics[scale=0.382]{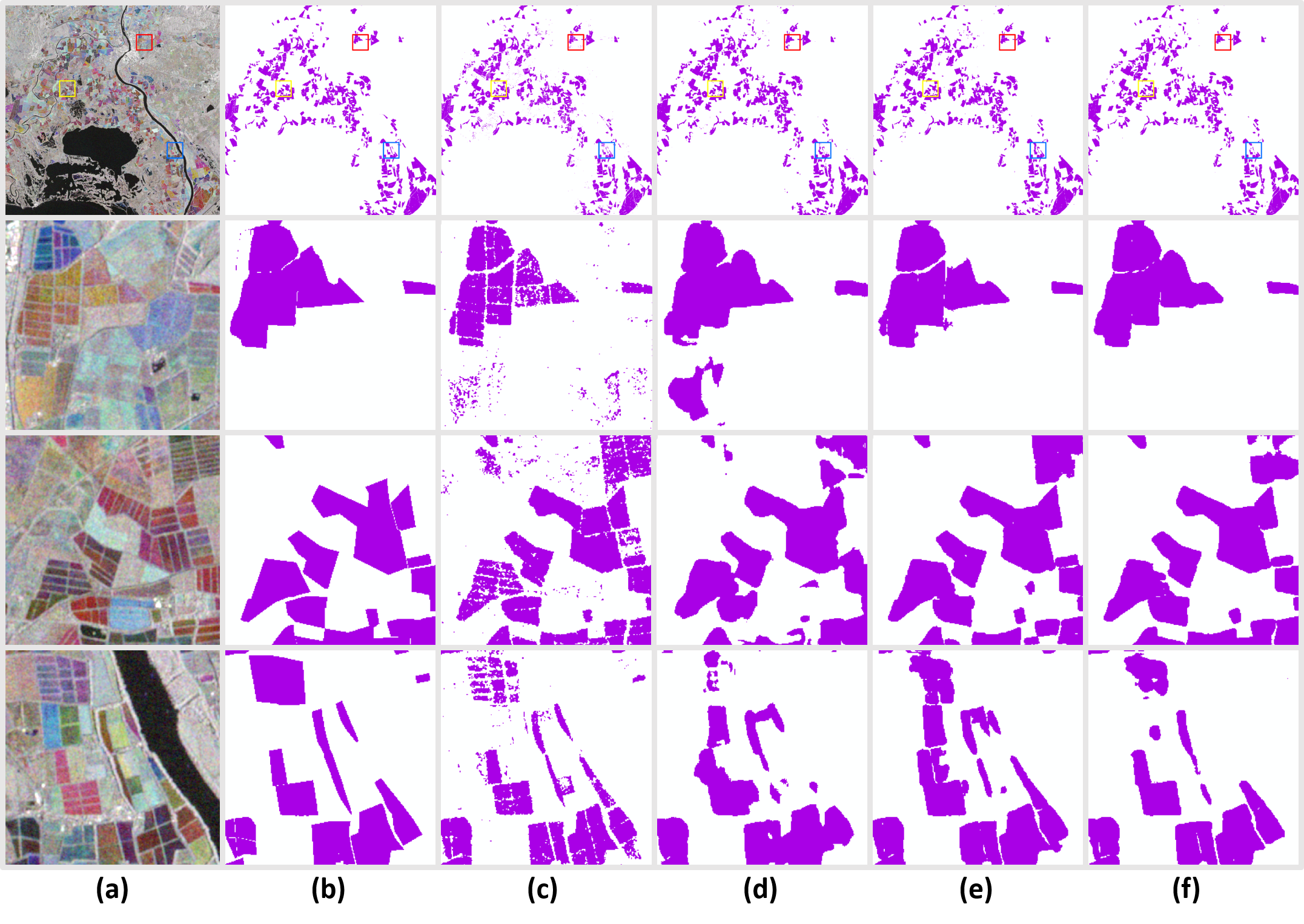}
    \caption{Visual comparisons on experiment 5 (France-France-rice-VH). The first row shows the overall results of the test image and the following rows show three randomly selected test patches. (a) Image. (b) Label. (c) RF. (d) RI. (e) FT. (f) FT\textsubscript{E}. }
    \label{fig:fra_fra_vh}
\end{figure}

\begin{figure}[!ht]
    \centering
    \includegraphics[scale=0.375]{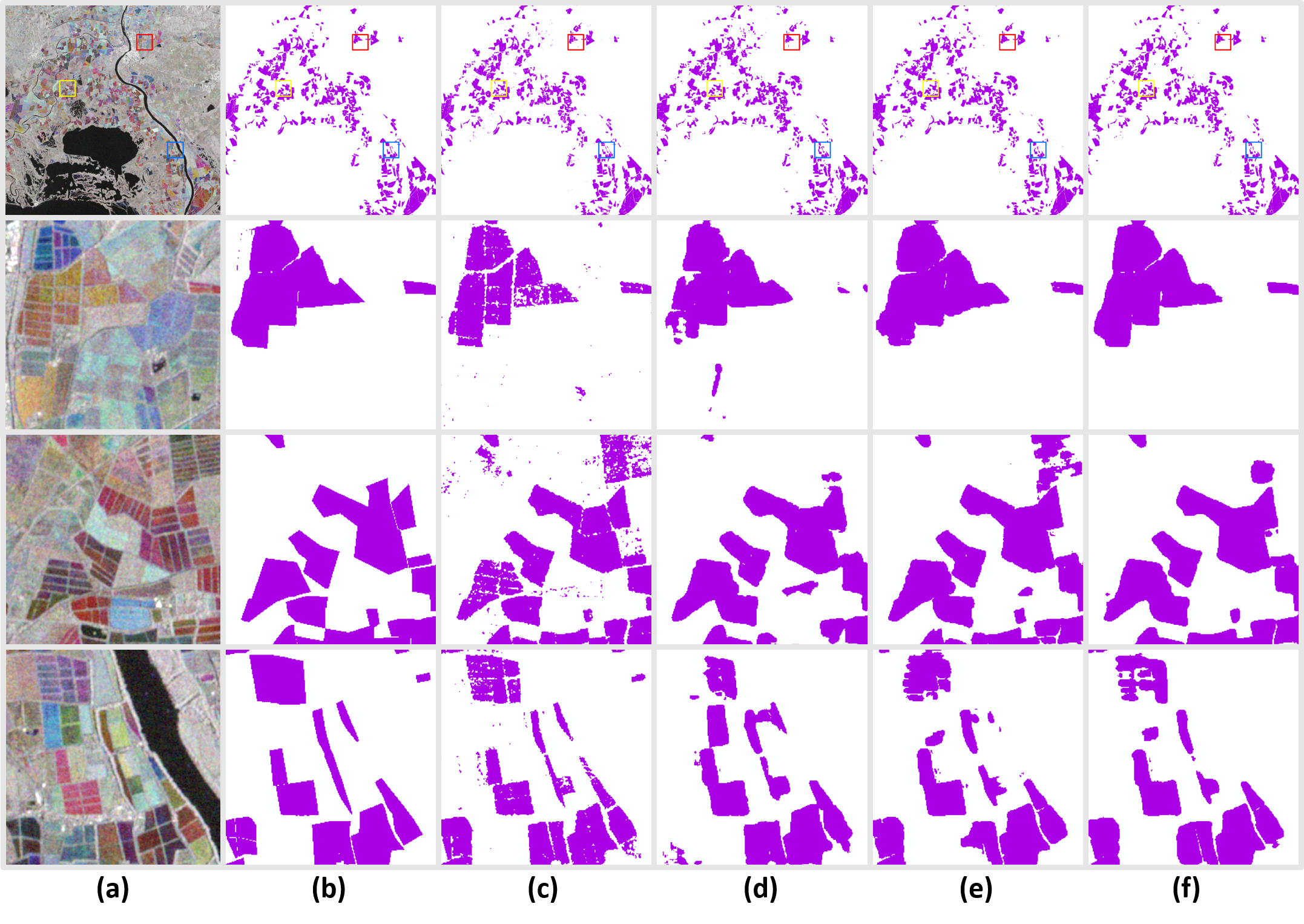}
    \caption{Visual comparisons on experiment 6 (France-France-rice-VH|VV). The first row shows the overall results of the test image and the following rows show three randomly selected test patches. (a) Image. (b) Label. (c) RF. (d) RI. (e) FT. (f) FT\textsubscript{E}. }
    \label{fig:fra_fra_vhvv}
\end{figure}

\begin{figure}[!ht]
    \centering
    \includegraphics[scale=0.375]{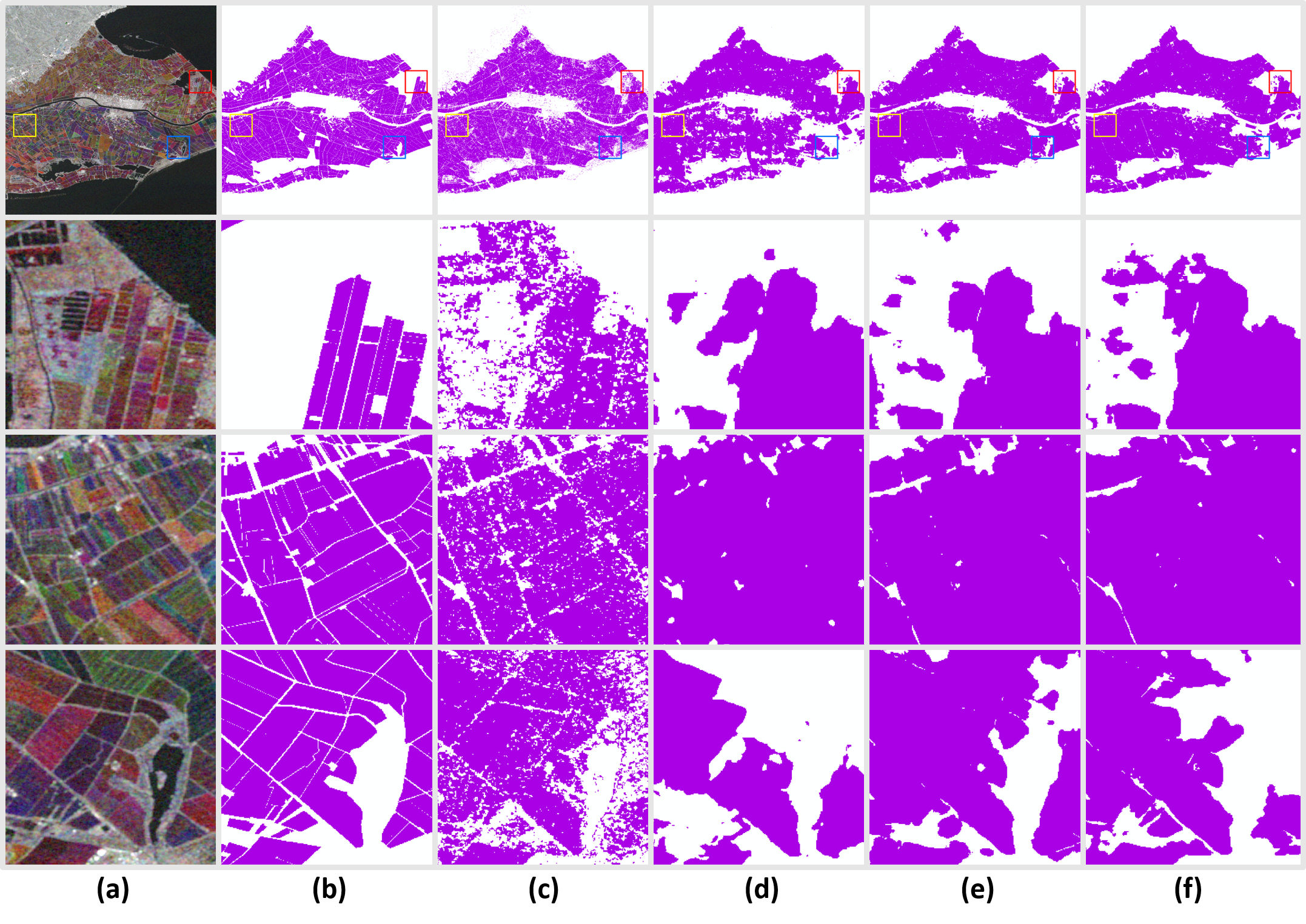}
    \caption{Visual comparisons on experiment 7 (France-Spain-rice-VH). The first row shows the overall results of the test image and the following rows show three randomly selected test patches. (a) Image. (b) Label. (c) RF. (d) RI. (e) FT. (f) FT\textsubscript{E}. }
    \label{fig:fra_spa_vh}
\end{figure}

\begin{figure}[!ht]
    \centering
    \includegraphics[scale=0.375]{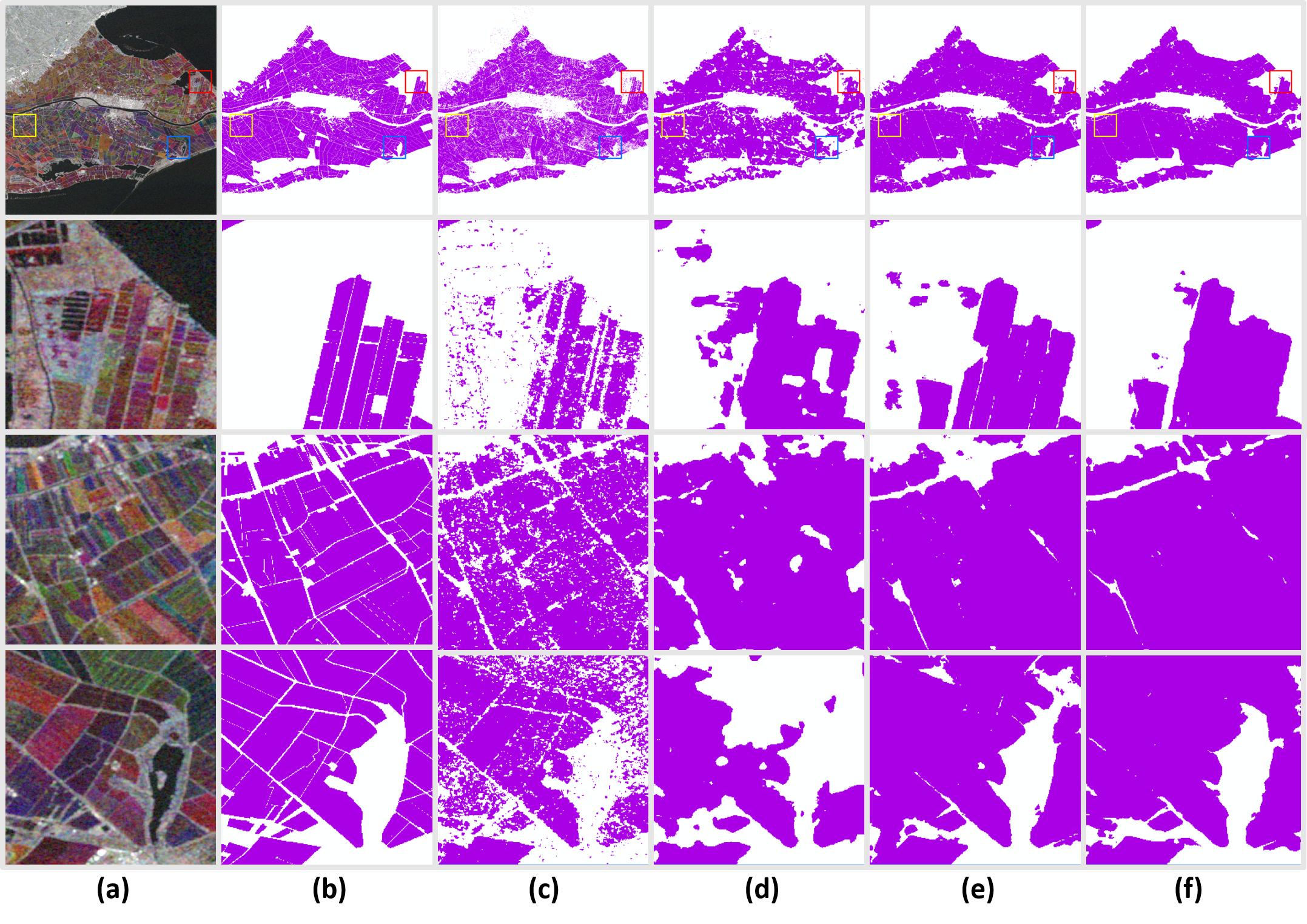}
    \caption{Visual comparisons on experiment 8 (France-Spain-rice-VH|VV). The first row shows the overall results of the test image and the following rows show three randomly selected test patches. (a) Image. (b) Label. (c) RF. (d) RI. (e) FT. (f) FT\textsubscript{E}. }
    \label{fig:fra_spa_vhvv}
\end{figure}

\begin{figure}[!ht]
    \centering
    \includegraphics[scale=0.375]{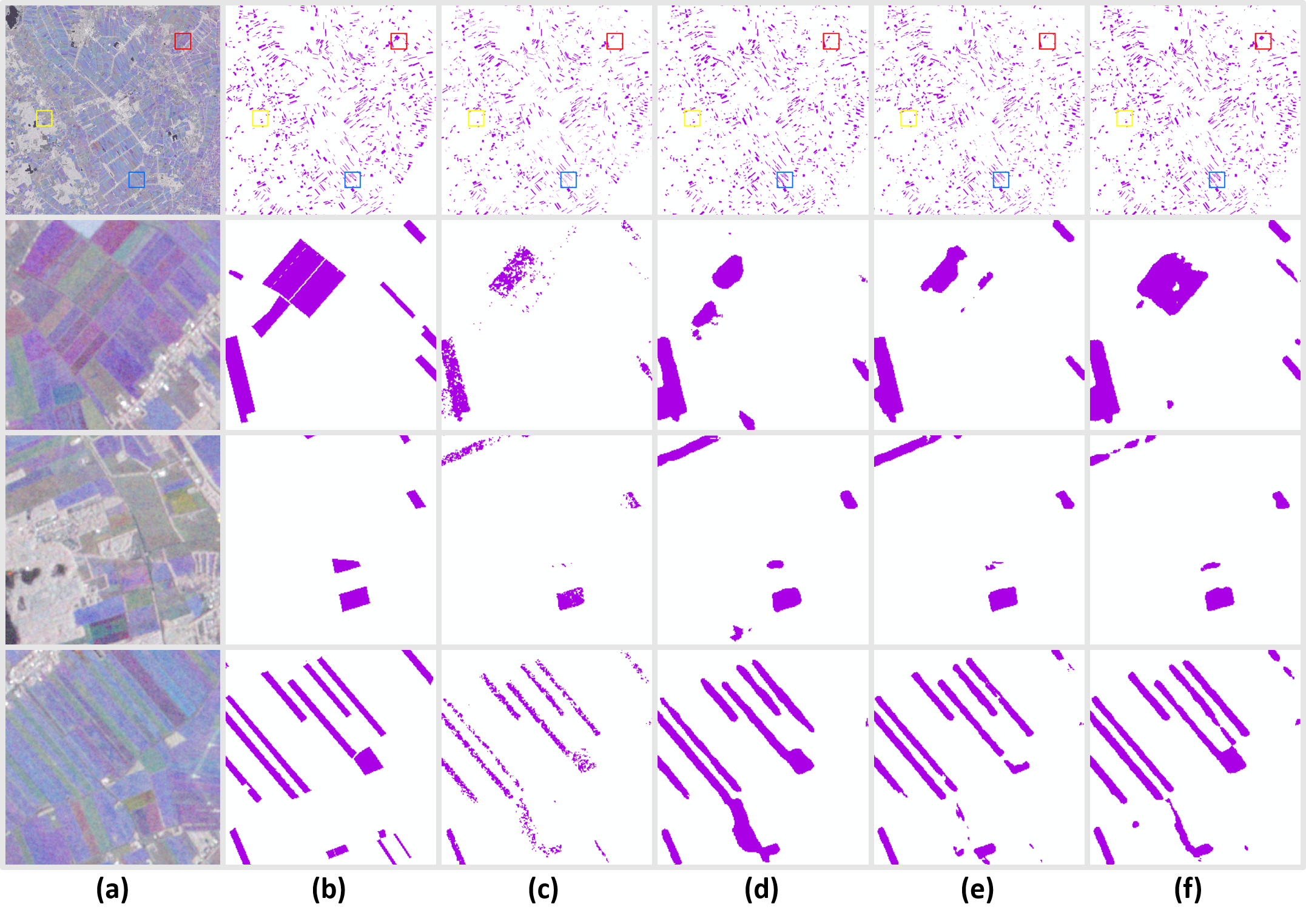}
    \caption{Visual comparisons on experiment 9 (Netherlands-Netherlands-barley-VH). The first row shows the overall results of the test image and the following rows show three randomly selected test patches. (a) Image. (b) Label. (c) RF. (d) RI. (e) FT. (f) FT\textsubscript{E}. }
    \label{fig:net_net_vh}
\end{figure}

\begin{figure}[!ht]
    \centering
    \includegraphics[scale=0.375]{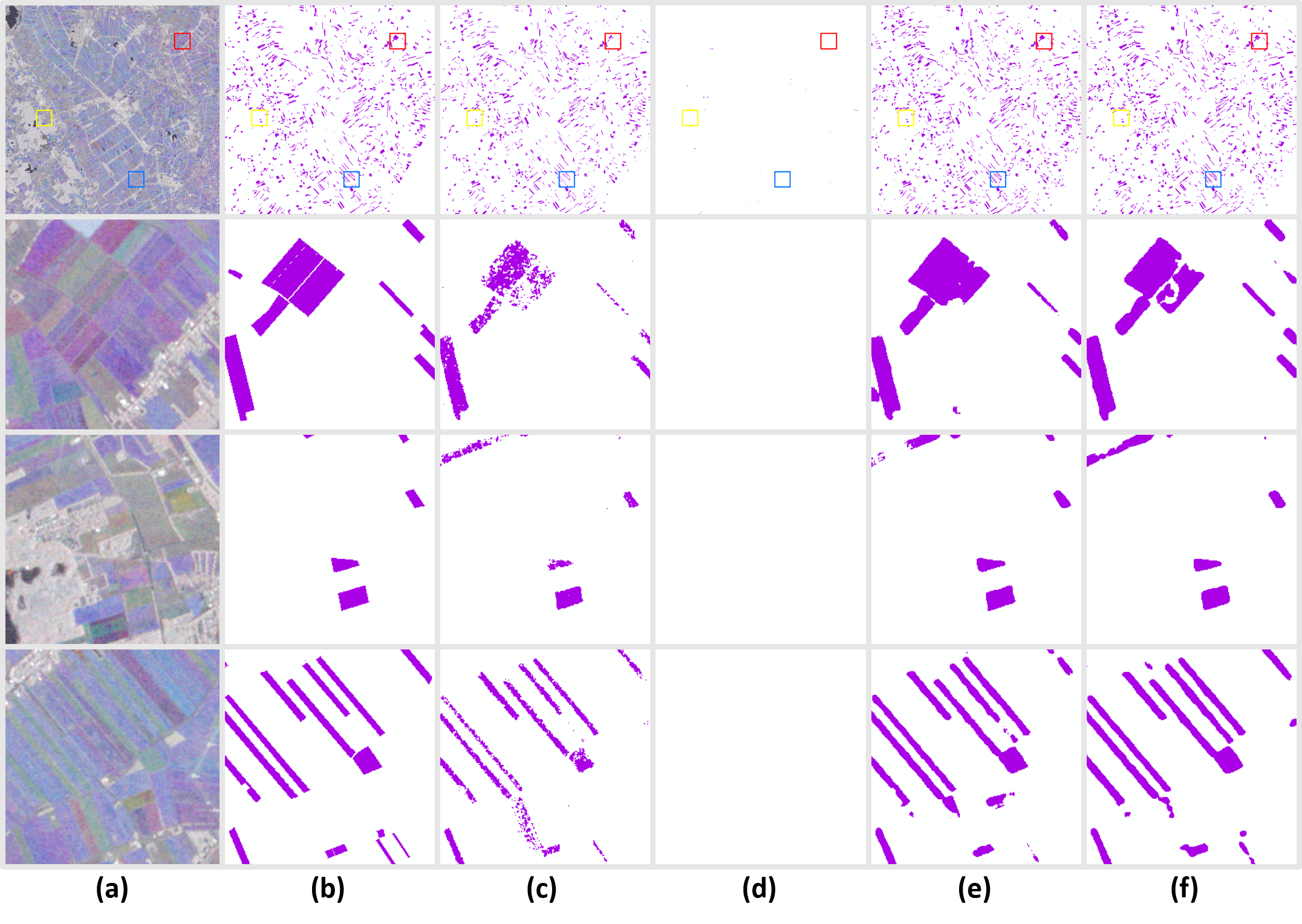}
    \caption{Visual comparisons on experiment 10 (Netherlands-Netherlands-barley-VH|VV). The first row shows the overall results of the test image and the following rows show three randomly selected test patches. (a) Image. (b) Label. (c) RF. (d) RI. (e) FT. (f) FT\textsubscript{E}. }
    \label{fig:net_net_vhvv}
\end{figure}
}
\end{document}